\documentclass[runningheads]{llncs}
\usepackage{graphicx}

\usepackage{tikz}
\usepackage{comment}
\usepackage{amsmath,amssymb}
\usepackage{color}

\usepackage{hyperref}
\usepackage{url}
\usepackage{wrapfig}
\usepackage{arydshln}
\usepackage{booktabs}
\usepackage{mathtools}
\usepackage{xcolor,colortbl}
\usepackage{blindtext}
\usepackage{caption}
\usepackage{multirow}
\usepackage{changepage,threeparttable} 
\usepackage{pifont}

\newcommand{\old}[1]{}
\newcommand{\rt}{\mathbin{:}}
\definecolor{lightest_gray}{gray}{0.95}
\usepackage{enumitem}
\newcommand{\etal}{\textit{et al. }}
\usepackage[usestackEOL]{stackengine}
\makeatletter
\newcommand{\mathleft}{\@fleqntrue\@mathmargin0pt}
\newcommand{\mathcenter}{\@fleqnfalse}
\makeatother
\usepackage[accsupp]{axessibility}

\begin{document}
\pagestyle{headings}
\mainmatter
\def\ECCVSubNumber{1551}

\title{Shift-tolerant Perceptual Similarity Metric}

\titlerunning{Shift-tolerant Perceptual Similarity Metric}
\author{Abhijay Ghildyal
\and
Feng Liu}
\institute{Portland State University, OR 97201, USA \\
\email{\{abhijay,fliu\}@pdx.edu}
}

\maketitle

\begin{abstract}
Existing perceptual similarity metrics assume an image and its reference are well aligned. As a result, these metrics are often sensitive to a small alignment error that is imperceptible to the human eyes. This paper studies the effect of small misalignment, specifically a small shift between the input and reference image, on existing metrics, and accordingly develops a shift-tolerant similarity metric. This paper builds upon LPIPS, a widely used learned perceptual similarity metric, and explores architectural design considerations to make it robust against imperceptible misalignment. Specifically, we study a wide spectrum of neural network elements, such as anti-aliasing filtering, pooling, striding, padding, and skip connection, and discuss their roles in making a robust metric. Based on our studies, we develop a new deep neural network-based perceptual similarity metric. Our experiments show that our metric is tolerant to imperceptible shifts while being consistent with the human similarity judgment. Code is available at \url{https://tinyurl.com/5n85r28r}.
\end{abstract}
\section{Introduction} \label{sec:intro}

Image similarity measurement is a common task for many computer vision and computer graphics applications. General similarity metrics like PSNR and RMSE, however, do not match the human visual perception well when assessing the similarity between two images. Therefore, many dedicated image similarity metrics, such as Structural Similarity (SSIM) and its variations~\cite{wang2004image,wang2003multiscale,zhang2011fsim,wang2005translation}, were developed in order to more closely reflect human perception. However, manually crafting a perceptual similarity metric remains a challenging task as it involves the complex human cognitive judgment~\cite{medin1993respects,tversky1977features,wang2004image,zhang2018perceptual}.

Recently, learning-based image similarity metrics have been developed, which learn from a large set of labeled data and predict similarity between images that correlates well with human perception~\cite{sangnie2020,Ding20,kettunen2019lpips,Prashnani_2018_CVPR,zhang2018perceptual,czolbe2020loss}. Among them, the Learned Perceptual Image Patch Similarity metric (LPIPS) by Zhang \etal, is a widely adopted metric and used in computer graphics and vision literature~\cite{zhang2018perceptual}.

This paper studies how similarity metrics work on a pair of images that are not perfectly aligned. For instance, a tiny one-pixel translation in the image pair, is imperceptible to humans. But, \emph{will such a visually imperceptible misalignment compromise any existing similarity metrics}? PSNR and RMSE assume pixel-wise registration, naturally making them sensitive even to a one-pixel misalignment. Our study found that learned metrics, such as LPIPS are also sensitive to a tiny\setlength{\intextsep}{-10pt}
\begin{wrapfigure}[14]{r}{0.48\textwidth}
\setlength{\abovecaptionskip}{4pt plus 0pt minus 0pt}
    \centering
    \scriptsize
    \includegraphics[width=0.47\textwidth]{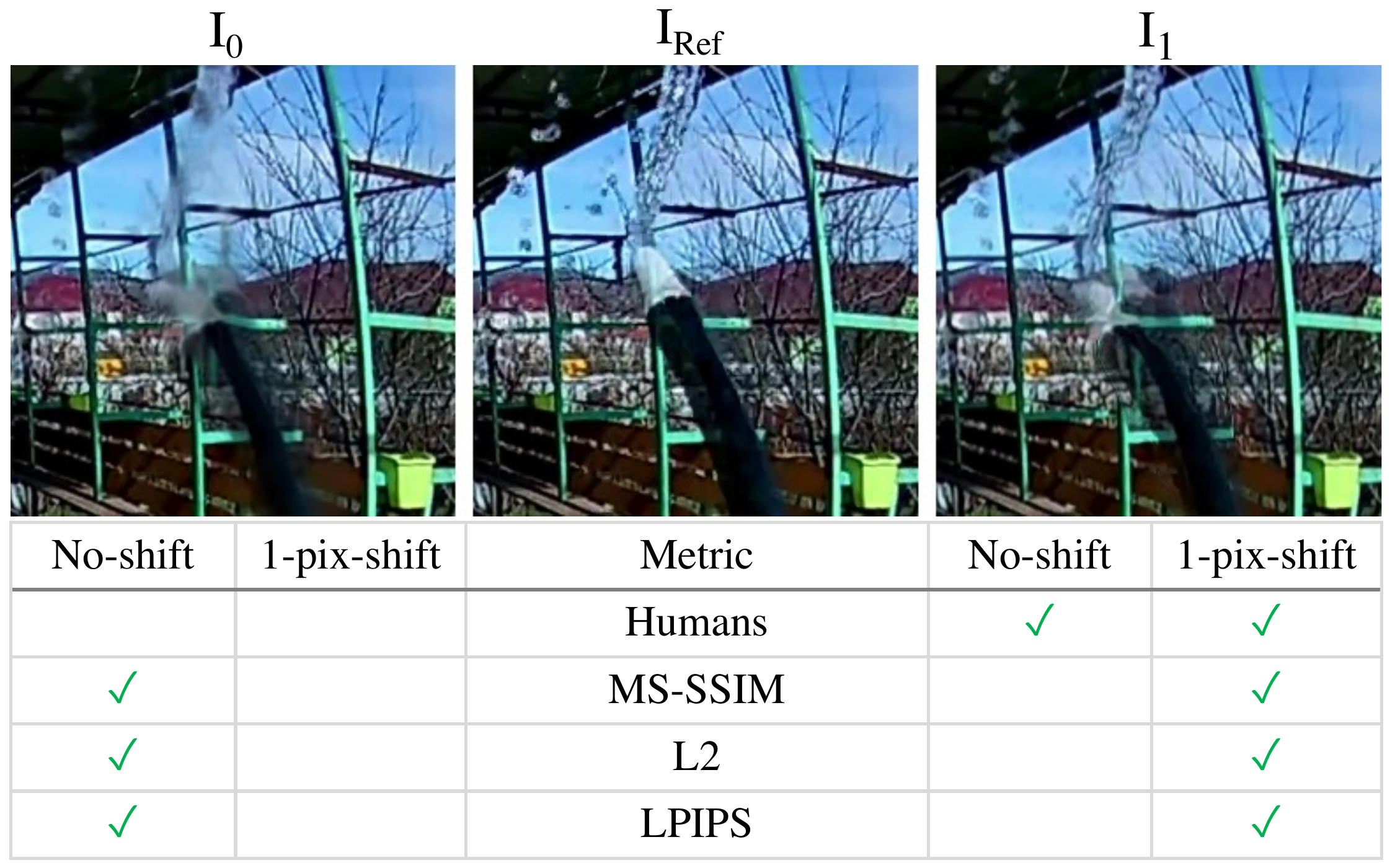}
    \caption{Whether $I_0$ and $I_1$ are shifted by 1 pixel or not, viewers always consider $I_1$ more similar to $I_{ref}$ than $I_0$; but existing similarity metrics often switch their predictions after the shift.}
\label{fig:teaser}
\end{wrapfigure}misalignment. Figure~\ref{fig:teaser} shows such an example via a two-alternative forced choice test, in which we asked viewers ``\emph{which of the two distorted images, $I_0$ or $I_1$, is more similar to the reference image $I_{ref}$}?" Then, we shifted $I_0$ and $I_1$ by one pixel and obtained opinions again. None of the participants flipped from $I_0$ to $I_1$ or vice versa, which is intuitive as a one-pixel shift is imperceptible to viewers. But existing metrics, such as MS-SSIM and LPIPS, flipped judgment after the one-pixel shift.

Our problem is related to the recent work on making deep neural networks shift invariant~\cite{Islam*2020How,Kayhan_2020_CVPR,vasconcelos2021impact,zhang2019shiftinvar,zou2020delving,leeJake}. Recently, Azulay and Weiss found that an image classifier can change its top-1 prediction if the image is translated by only one pixel~\cite{azulay2018deep}. Their results showed that after translating an image by one pixel, the classifier made a different top-1 prediction for 30\% of the 1000 validation images. Zhang introduced anti-aliasing filters into a deep neural network to make the feature extraction network shift-equivariant, which in terms makes the whole network shift-invariant for the down streaming tasks~\cite{zhang2019shiftinvar}. Compared to these works, our problem is different in that 1) a perceptual similarity metric takes two images as input instead of working on a single input image, and 2) only one of the two images is shifted, thus introducing imperceptible misalignment instead of shifting the two images simultaneously.

This paper aims to develop a shift-tolerant perceptual similarity metric that correlates well with the human judgment on the similarity between images while being robust against imperceptible misalignment between them. We build our metric upon LPIPS, a deep neural network-based metric widely adopted for its close correlation with human perception. We investigate a variety of elements that can be incorporated into a deep neural network to make it resistant to an imperceptible misalignment. These elements include anti-aliasing filters, striding, pooling, padding, placement of anti-aliasing, etc. Based on our findings on these elements, we develop a shift-tolerant perceptual similarity metric that not only is more consistent with human perception but also is significantly more resistant to imperceptible misalignment between a pair of images than existing metrics.

In the remainder of this paper, we first report our study, verifying that viewers are not sensitive to small amounts of shifts between two images when comparing them, in Section~\ref{sec:user_study}. We then benchmark existing similarity metrics and show that they are sensitive to imperceptible shifts between a pair of images in Section~\ref{sec:benchmark}. We study several important elements that make a deep neural network-based similarity metric both tolerant to imperceptible shifts and consistent with human perception of visual similarity in Section~\ref{sec:methods}. We finally report our experiments that thoroughly evaluate our new perceptual similarity metric by comparing it to state-of-the-art metrics and through detailed ablation studies in Section~\ref{sec:experiments}.

\section{Related Work} \label{sec:related_work}

Visual similarity metrics are commonly used to compare two images or evaluate the performance of many image and video processing, editing and synthesis algorithms. While there are already many established metrics for these tasks, such as PSNR, MSE, SSIM and its variations~\cite{wang2004image,wang2003multiscale,wang2005translation}, there is still a gap between their prediction and the human's judgment. This section provides a brief overview of the recent advances in learned perceptual similarity metrics that aim to bridge the gap mentioned above. 

In their influential work, Zhang \etal reported that features from a deep neural network can be used to measure the similarity between two images that is more consistent with the human perception than other commonly used metrics~\cite{zhang2018perceptual}. Accordingly, they developed LPIPS, a perceptual metric learned from a large collection of labelled data. Specifically, LPIPS uses a pre-trained network for image classification tasks or learns a neural network to compute the features for each of the two images or patches, and also learns to aggregate the feature distances into a similarity score. Since its debut, LPIPS has been widely used as a perceptual quality metric. On a related note, the computer vision and graphics community also calculate the difference between the deep features of two images as a loss function to train deep neural networks for image enhancement and synthesis. Such a loss function, often called perceptual loss, enables the neural networks to learn to generate perceptually pleasing images~\cite{Dosovitskiy_NIPS_2016,Johnson_ECCV_2016,Ledig_CORR_2016,niklaus2017video,Sajjadi_CORR_2016,Zhu_ECCV_2016}.

Kettunen \etal developed the E-LPIPS metric that adopts the LPIPS network and uses randomly transformed samples to calculate expected LPIPS distance over them~\cite{kettunen2019lpips}. They showed that E-LPIPS is robust against the Expectation Over Transformation attack~\cite{pmlr-v80-athalye18a}. Different from LPIPS, Prashnani \etal use the differences between features to generate patch-wise errors and corresponding weights, via two different fully-connected networks~\cite{Prashnani_2018_CVPR}. Their final similarity score is a weighted average of the patch-wise distances. Czolbe \etal developed a similarity metric based on Watson's perceptual model~\cite{watson1993dct}, by replacing discrete cosine transform with discrete fourier transform~\cite{czolbe2020loss}. They posit that their metric is robust against small translations and is sensitive to large translations. 

In an earlier work, Wang \& Simoncelli~\cite{wang2005translation} improved SSIM~\cite{wang2004image} by replacing the spatial correlation measures with phase correlations in wavelet subbands which made the metric less sensitive to geometric transformations. Ma~\etal \cite{Ma18:geo} developed a geometric transformation invariant method (GTI-CNN). Our work is closely related to theirs, as GTI-CNN is a similarity metric invariant to the misalignment between a pair of images. In their method, Ma \etal train a fully convolutional neural network to extract deep features from each image and calculate the mean squared error between them as final similarity~\cite{Ma18:geo}. They showed that training the network directly on aligned samples leads to a metric sensitive to misalignment, which is consistent with what we found in our study. They reported that augmenting the training samples with small misalignment can make the learned metric significantly more resistant to the misalignment. Compared to this method, our work focuses on designing a deep neural network architecture robust to misalignment without any data augmentation. Bhardwaj \etal followed the understanding of the physiology of the human visual system and developed a fully convolutional neural network that generates a multi-scale probabilistic representation of an input image and then calculates the symmetric Kullback–Leibler divergences between such representations of two images to measure their similarity \cite{sangnie2020}. They found their similarity metric, perceptual information metric (PIM), robust against small shifts between a pair of images. While benchmarking existing metrics, our study also finds that PIM is most robust against small shifts among all metrics tested. We posit that the robustness of PIM partially comes from training on neighboring video frames that might already have small shifts among them, thus effectively serving as data augmentation, as done by Ma \etal \cite{Ma18:geo}. We consider these as orthogonal efforts in developing a robust metric. As shown in our study, our metric is more consistent with human judgment and more robust against imperceptible misalignment than these methods, except PIM, to which ours is comparable, although our metric is trained on aligned samples directly without any data augmentation. 

Our work is most related to deep image structure and texture similarity (DISTS) metric by Ding \etal \cite{Ding20}. They used global feature aggregation to make DISTS robust against mild geometric transformations. They also replaced the max pooling layers with $l_2$ pooling layers~\cite{henaff2016geodesics} in their VGG backbone network for anti-aliasing and found that blurring the input with $l_2$ pooling makes their network more robust against small shifts. Gu \etal \cite{gu_2020_eccv} found that existing metrics like LPIPS do not perform well with images generated by GAN-based restoration algorithms. They attributed it to the small misalignment between the GAN results and the ground truth. Therefore, they used  $l_2$ pooling~\cite{Ding20,henaff2016geodesics} and \emph{BlurPool}~\cite{zhang2019shiftinvar} to improve LPIPS. They found that both can improve LPIPS while \emph{BlurPool} performs better. Compared to these two recent papers, our paper systematically investigates a broad range of neural network elements besides \emph{BlurPool}. By integrating these elements, we develop a perceptual similarity metric both robust against small shifts and consistent with the human visual similarity judgment. Our method outperforms existing metrics and a variety of recently developed learned metrics. Integrating multiple network elements together makes our metric better than individual ones.

\section{Human Perception of Small Shifts} \label{sec:user_study}

As commonly expected, shifting one image by a few pixels will not alter human similarity judgment on a pair of images~\cite{sangnie2020,xiao2018spatially}. We conducted a user study to verify this common belief. Our hypothesis is that \emph{it is difficult for people to detect a small shift in images}. In our study, we randomly chose 50 images from the MS-COCO test dataset~\cite{lin2014microsoft} and divided them into 10 groups, each with 5 images. For each image in Group $n$ with  $n \in [0, 10)$, we cropped a $256 \times 256$ patch as a reference image and shifted the cropping window by $n$ pixels to produce its shifted version. Since we always cropped the reference from the same location, we had an n-pixel shifted version for each of the 50 images, and thus in total, we have 500 pairs of images in our study. For each participant, we randomly sampled 50 pairs from the 500, with 5 pairs for each 0-9 pix-shift. The 50 pairs were presented one at a time, with the two images placed side by side. The position of the reference image, right, or left, is randomized to avoid biases. We asked our participants to judge whether a pair of images are the same or not.

To maintain the quality of our study and avoid boring the users, we only presented 50 samples to each user. Interestingly, humans managed to detect the shift for a 2 pixel shift in ~50\% of cases. We attribute this partially to the fact that the users were informed that there might or might not be a shift between a pair of images. While this might bias participants, such that their sensitivity to the shifts is likely increased, we found it helpful to obtain a more informative understanding of the human perception of small shifts; otherwise, participants tended to overly overlook the difference between a pair of images. Specifically, in our pilot study, we found that users were very confused when we asked them if a pair of images looked the same or not. Many of them thought if we were asking them to compare high-level features such as objects in the two images or if there were some artifacts in one of the pair of images.

We recruited 32 participants for our study. They have a wide range of professional backgrounds, including computer science, business, medicine, arts, and education, and most of them are 20 to 35 years old. To ensure the quality of this user study, we removed the responses from two participants who failed to pass a validation test. If a participant identified a pair of images with 0-pixel shift as \emph{different} or a pair of images with 9-pixel shift as \emph{the same} for more than two-thirds of the time in the study, we exclude all responses from that participant. In total, 30 participants passed our validation test. We obtained responses to 1500 trials in total, with 150 responses for each of the $n$-pixel shifts.

Finally, we report the user responses in Table~\ref{tab:pixelShifts}. When the amount of shift is\setlength{\intextsep}{-3pt}
\begin{wraptable}[19]{r}{0.48\textwidth}
\setlength{\abovecaptionskip}{7pt plus 2pt minus 2pt}
    \begin{center}
    \scriptsize
    \caption{Human perception of small shifts. Image pairs with 1- and 2-pixel shift are deemed the same in 80.7\% and 56.0\% of the responses, resp. The avg. of std. in responses per sample indicates that users were more doubtful about image pairs with 2-5 pixel shift.}
    \label{tab:pixelShifts}
    \begin{tabular}{ccccc}\toprule
    
    \multirow{3}{*}{\Centerstack{Pixel \\ shift}}
    & \multicolumn{3}{c}{Number of user responses}
    & \multirow{4}{*}{\Centerstack{Avg. of std. in \\ user responses \\ per sample}} \\
    \cmidrule(lr){2-4}
    
    & Said Yes
    & Said No
    & Yes\%
    & \\
    
    & (Same)
    & (Shifted)
    & 
    &  \\
    \midrule
    0 & 140 & 10 & 93.3\% & 0.09 $\pm$ 0.17 \\
    1 & 121 & 29 & 80.7\% & 0.19 $\pm$ 0.23 \\
    2 & 84 & 66 & 56.0\% & 0.34 $\pm$ 0.21 \\
    3 & 52 & 98 & 34.7\% & 0.24 $\pm$ 0.23 \\
    4 & 52 & 98 & 34.7\% & 0.30 $\pm$ 0.24 \\
    5 & 40 & 110 & 26.7\% & 0.23 $\pm$ 0.24 \\
    6 & 35 & 115 & 23.3\% & 0.21 $\pm$ 0.24 \\
    7 & 31 & 119 & 20.7\% & 0.12 $\pm$ 0.20 \\
    8 & 27 & 123 & 18.0\% & 0.18 $\pm$ 0.23 \\
    9 & 15 & 135 & 10.0\% & 0.13 $\pm$ 0.21 \\
    \bottomrule
    \end{tabular}
\end{center}
\end{wraptable}small, participants find it difficult to detect the shift. Samples with 1- and 2-pixel shifts were considered the same in 80.7\% and 56.0\% of the responses, respectively. As expected, shifts become easier to detect as the size of shifts increases. But even pairs with a 5-pixel shift were still not identified in 26.7\% of the responses. We further analyzed the variability in user responses per sample, grouping them by the amount of pixel-shift. As observed in Table~\ref{tab:pixelShifts}, with no or only a 1-pixel shift, users were consistently sure that the images in each pair were the same. Similarly, users could consistently detect the large shift (6 to 9 pixels). However, for a 2 to 5-pixel shift, we observed that users were doubtful whether images were shifted or not, indicated by the high variation in responses. As shown in our study, even after being informed about the possible shifts, participants still had difficulty detecting small shifts. This verifies our hypothesis that it is difficult for people to detect a small shift in images. In addition, we use this data and test the consistency of various metrics with the sensitivity of human perception to pixel shifts in Section~\ref{sec:experiments}. The test results provide further evidence that our metric is more consistent with human perception.
    
\section{Effect of Small Shifts on Similarity Metrics} \label{sec:benchmark}

To understand how existing similarity metrics handle small shifts between a pair of images, we benchmarked representative metrics, including off-the-shelf metrics, such as L2 and SSIM, and the recent deep learning-based metrics. We derived a new dataset from the Berkeley-Adobe Perceptual Patch Similarity Dataset (BAPPS)~\cite{zhang2018perceptual}. The original BAPPS data consists of 36,344 samples, each with a reference $I_r$, and two distorted images $I_1$ and $I_2$. They cover a wide range of common distortions, including traditional, CNN-based, and from algorithms such as superresolution, frame interpolation, deblurring, and colorization. Please refer to \cite{zhang2018perceptual} for more details. For each sample in the BAPPS dataset, we shifted the distorted images horizontally by $k$ pixels where $k \in \{1, 2, 3\}$. To avoid any boundary artifacts from shifting, we cropped each shifted image $I_i$ as follows.
\begin{equation}
    \label{eq:cropping}
    \hat{I}_i = I_i[\; 0 \rt h, \; k \rt (w + k - 3 )]
\end{equation}
where $w$ and $h$ are the original sizes. In this way, all the images in our test were of size $(w-3) \times h$ without regard to the amount of shift, which eliminates the effect\begin{wraptable}[20]{r}{0.5\textwidth}
\setlength{\abovecaptionskip}{7pt plus 2pt minus 2pt}
    \begin{center}
    \scriptsize
    \caption{Accuracy (2AFC) and shift-tolerance ($r_{rf}$) of various metrics on the BAPPS val. dataset. 2AFC score is computed on the BAPPS data resized to $64 \times 64$ while $r_{rf}$ scores are obtained from its shifted version of size $64 \times 61$.}
    \label{tab:ti_study64}
    \begin{tabular}{lcccc}\toprule
        \multirow{3}{*}{Network} &\multirow{3}{*}{2AFC} &\multicolumn{3}{c}{$r_{rf}$} \\\cmidrule(lr){3-5}
        & &1pixel &2pixel &3pixel \\
        \midrule  
        L2 &62.91 &12.27 &23.07 &28.83 \\
        SSIM~\cite{wang2004image} &63.08 &13.08 &25.50 &32.74 \\
        CW-SSIM~\cite{wang2005translation} &60.55 &11.33 &18.28 &23.22 \\
        E-LPIPS~\cite{kettunen2019lpips} &69.23 &8.72 &10.67 &12.34 \\
        GTI-CNN~\cite{Ma18:geo} &63.74 &9.37 &12.32 &16.25 \\
        DISTS~\cite{Ding20} &68.89 &5.57 &8.20 &10.07 \\
        PIM-1~\cite{sangnie2020} &69.45 &\textbf{1.63} &\textbf{3.06} &\textbf{4.39}\\
        PIM-5~\cite{sangnie2020} &69.47 &2.28 &3.56 &5.19 \\
        LPIPS (Alex)~\cite{zhang2018perceptual} &69.83 &6.79 &8.90 &9.70 \\
        LPIPS (Alex) $^{\S * \dag}$ &\textbf{70.04} &9.25 &9.34 &11.55 \\
        LPIPS (Alex) ours$^{* \dag}$ &69.83 &3.48 &4.75 &6.84 \\
        \bottomrule
    \end{tabular}
    \end{center}
    \scriptsize{($\S$) Retrained from scratch. ($*$) Trained on image patches of size 64 using author's ($\dag$) setup.}
\end{wraptable}of image sizes when we test how the amount of shift affects the performance. The references are cropped to the same size as the distorted images, but no shifts were applied. In addition, we also cropped all the images in each original sample to the size of $(w-3) \times h$ to make the shifted sample and the original sample the same size to avoid the effect of the image size on a similarity metric in our late experiments. No shift was introduced to the original samples. A 3-pixel shift in our setting is equivalent to shifting 1.2\% of the pixels for the images of size $256 \times 256$ pixels.

When evaluating a metric, we apply it to both the original sample from BAPPS and its corresponding shift, i.e., for each sample, we obtain two pairs of similarity scores, $(s_1, s_2)$ and $(\hat{s}_1, \hat{s}_2)$. $(s_1, s_2)$ are the similarity scores between $I_1$ and its reference $I_r$, and $I_2$ and $I_r$, respectively. $(\hat{s}_1, \hat{s}_2)$ are the corresponding pair of similarity scores for the shifted sample. Each pair of scores indicates which of the two distorted images is more similar to the reference according to the metric. We count the number of samples for which the similarity rank flipped when a sample was shifted and compute the rank-flip rate as follows.
\begin{equation}
    \label{eq:2}
    r_{rf} = \frac{1}{N} \sum_{l=1}^{N} ( s_1^l < s_2^l) \neq ( \hat{s}_1^l < \hat{s}_2^l)
\end{equation}where  $r_{rf}$ is the rank-flip rate and $N$ is the number of samples. $r_{rf}$ evaluates how robust a metric is against the small shift between a pair of images.

For all learned metrics involved in this study, we used the trained models shared by their authors. While the image size in BAPPS is $256 \times 256$, some models were trained on $64 \times 64$ resized images. Thus, we conducted studies on these two sizes separately to provide fair comparisons. We report the results in Tables~\ref{tab:ti_study64} and ~\ref{tab:ti_study}. All scores are obtained by averaging over examples in each distortion category in BAPPS and then averaging over all the categories. The two-alternative forced choice (2AFC) scores were obtained from the original BAPPS dataset that indicates how a metric's prediction correlates with the human opinion~\cite{zhang2018perceptual}. The rank-flip rate ($r_{rf}$) is calculated from the shifted dataset. It shows how robust a metric is to the shift between a distorted image and its reference. As reported in Table~\ref{tab:ti_study}, learned metrics match human perception better than non-learned\begin{wraptable}[23]{r}{0.5\textwidth}
\setlength{\abovecaptionskip}{7pt plus 0pt minus 0pt}
\setlength{\belowcaptionskip}{-2.5pt plus 0pt minus 0pt}
    \begin{center}
    \scriptsize
    \caption{Accuracy (2AFC) and shift-tolerance ($r_{rf}$) across various metrics on the BAPPS val. dataset. 2AFC score is computed on the original BAPPS data of size $256 \times 256$ while $r_{rf}$ is obtained from its shifted version of size $256 \times 253$.}
    \label{tab:ti_study}
    \begin{tabular}{lcccc}\toprule
        \multirow{3}{*}{Network} &\multirow{3}{*}{2AFC} &\multicolumn{3}{c}{$r_{rf}$} \\\cmidrule(lr){3-5}
        & &1pixel &2pixel &3pixel \\
        \midrule 
        L2 &62.92 &3.59 &7.55 &10.82 \\
        SSIM~\cite{wang2004image} &61.41 &3.16 &7.20 &13.73 \\
        CW-SSIM~\cite{wang2005translation} &61.48 &3.91 &6.88 &9.47 \\
        MS-SSIM~\cite{wang2003multiscale} &62.54 &2.22 &5.83 &10.66 \\
        PIEAPP Sparse~\cite{Prashnani_2018_CVPR} &64.20 &2.83 &3.19 &3.81 \\
        PIEAPP Dense~\cite{Prashnani_2018_CVPR} &64.15 &2.97 &1.37 &3.33 \\
        PIM-1~\cite{sangnie2020} &67.45 &0.79 &1.70 &2.52 \\
        PIM-5~\cite{sangnie2020} &67.38 &1.01 &1.88 &2.96 \\
        GTI-CNN~\cite{Ma18:geo} &63.87 &3.95 &4.91 &7.88 \\
        DISTS~\cite{Ding20} &68.83 &2.85 &2.89 &4.03 \\
        E-LPIPS~\cite{kettunen2019lpips} &68.22 &5.84 &5.86 &5.77 \\
        LPIPS (Alex)~\cite{zhang2018perceptual} &68.59 &2.81 &3.41 &3.84 \\
        LPIPS (Alex) $^{\S * \dag}$ &70.54 &2.58 &3.59 &3.53 \\
        LPIPS (Alex) ours$^{* \dag}$ &70.39 &0.66 &1.24 &1.79 \\
        LPIPS (Alex) $^{\S * \ddag}$ &\textbf{70.65} &2.87 &3.92 &3.74 \\
        LPIPS (Alex) ours$^{* \ddag}$ &70.48 &\textbf{0.57} &\textbf{1.06} &\textbf{1.50} \\
        \bottomrule
    \end{tabular}
    \end{center}
    \scriptsize{($\S$) Retrained from scratch. ($*$) Trained on patches of size 256 using author's ($\dag$) / our ($\ddag$) setup.}
\end{wraptable}ones such as L2, SSIM, CW-SSIM and MS-SSIM. However, even these learned metrics are sensitive to small shifts except for the recent PIM~\cite{sangnie2020}. Compared to these existing metrics except PIM, our metrics are more consistent with human perception as per 2AFC and more robust against small shifts. Overall, our method is comparable to PIM. Our method outperforms PIM on images of size $256 \times 256$ (Table~\ref{tab:ti_study}) but does not work as well as it on smaller images (Table~\ref{tab:ti_study64}). As discussed in Section~\ref{sec:related_work}, PIM is trained on neighboring video frames that often contain small shifts, which makes it robust against the imperceptible shifts. Our work is orthogonal to PIM as we investigate neural network elements to build a robust similarity metric. Hence, we purposely trained our metrics on the BAPPS dataset without any data augmentation.

\section{Elements of Shift-tolerant Metrics} \label{sec:methods}

\setlength{\intextsep}{2pt}
\begin{wrapfigure}[18]{r}{0.5\textwidth}
\setlength{\abovecaptionskip}{7pt plus 2pt minus 2pt}
\scriptsize
    \centering
        \includegraphics[width=0.5\textwidth]{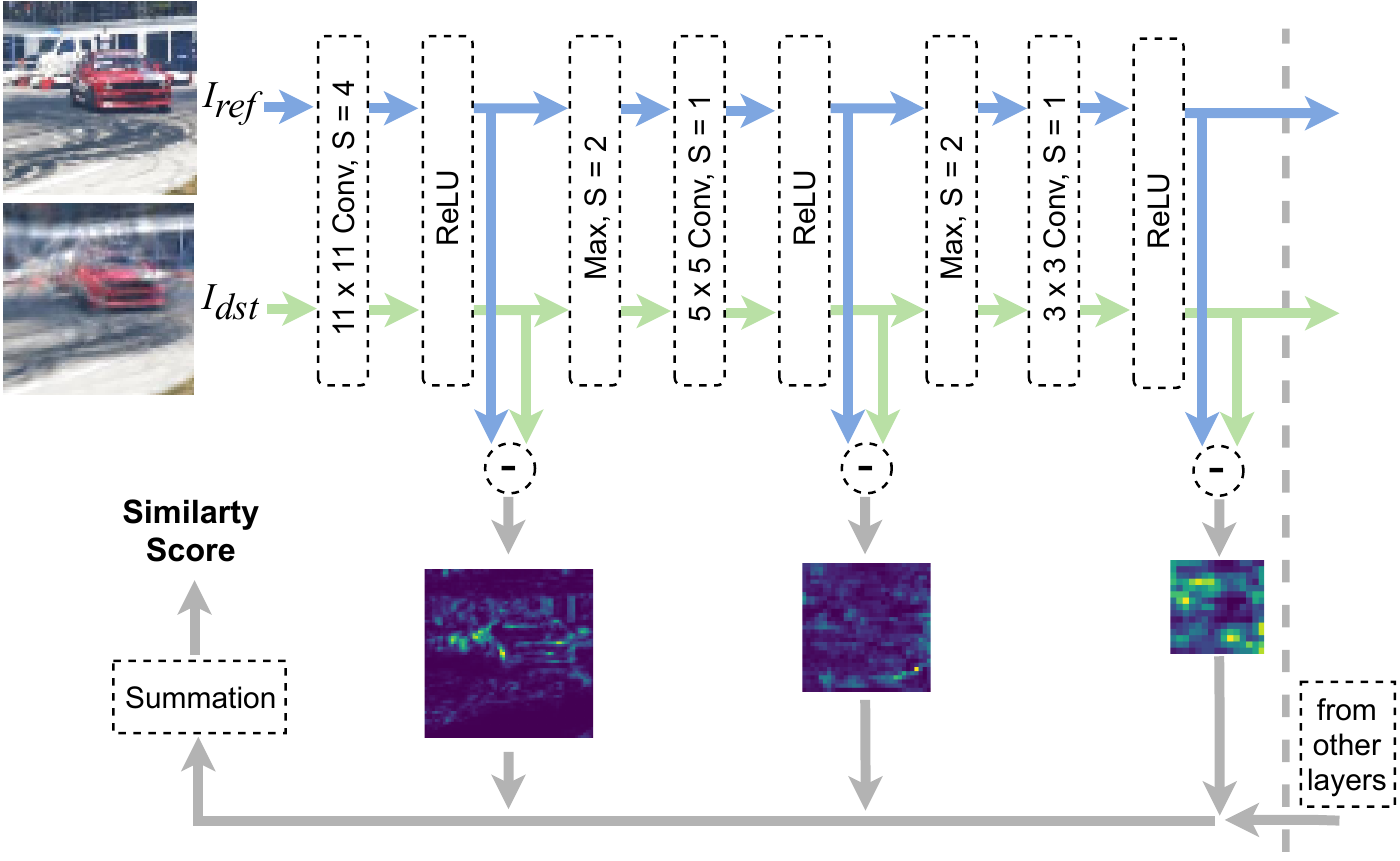}
    \caption{LPIPS framework. The same feature extraction network (AlexNet) is used to extract feature embeddings from  $I_{dst}$ and $I_{ref}$. The difference between these embeddings is calculated at different levels and is combined together as the similarity between $I_{dst}$ and $I_{ref}$.}
\label{fig:lpips_arch}
\end{wrapfigure}
Some recent papers reported that training a deep neural network using samples with shifted images through either data augmentation or neighboring video frames can make a learned similarity metric robust against small shifts between a pair of images~\cite{Ma18:geo,sangnie2020}. This paper aims to solve this problem from a different perspective; we investigate how one can design a deep neural network resistant to small shifts. We select the LPIPS network architecture as our baseline framework as it correlates with the human visual similarity judgment well~\cite{zhang2018perceptual}. To make this paper self-complete, we briefly describe the LPIPS framework. As illustrated in Figure~\ref{fig:lpips_arch}, LPIPS uses a backbone network, such as AlexNet~\cite{Krizhevsky_NIPS_2012} or VGG~\cite{simonyan2014very}, to extract multi-level feature embeddings from a distorted image $I_{dst}$ or its reference image $I_{ref}$. We denote the resulting feature embeddings as $F_{dst}$ and $F_{ref}$, respectively. It then calculates the difference between $F_{dst}$ and $F_{ref}$ at all the levels and linearly combines the embedding difference at different levels into a final similarity / difference score, denoted as $d(F_{dst}, F_{ref})$. The combination coefficients and the feature extraction network are learned or fine-tuned.

Below we discuss how various neural network elements affect a similarity metric and how they can be improved to handle imperceptible shifts between a pair of images. Our focus is to develop a feature extraction network to generate feature embeddings from a pair of images that 1) are invariant to imperceptible shifts and 2) lead to a metric that correlates well with the human judgments. 
    
\textbf{Reducing Stride.} Striding is widely used in a deep neural network to reduce the input size. For instance, AlexNet has a strided convolution (stride=4) in its first convolutional layer (\emph{conv-1}) and many max pooling operators with stride=2 in the rest of the network. However, it is commonly known that striding with size $>$1 leads to the sampling rate falling well below the Nyquist rate, which causes aliasing artifacts. In their experiments with image classification tasks, \cite{azulay2018deep} showed that AlexNet without any subsampling is significantly less sensitive to translations and also maintains its accuracy. Similarly, we also investigate the reduction of the stride size in the convolutional layers in the LPIPS framework to make it more resistant to imperceptible shifts at no expense of its consistency with the human visual similarity perception. 

\textbf{Anti-aliasing.} Convolution is the most common operator for a deep convolutional neural network. A pure convolutional operator is shift-equivariant instead of being shift-invariant~\cite{nair2010rectified}. Shift equivariance makes a learned similarity metric\setlength{\intextsep}{2pt}
\begin{wrapfigure}[17]{r}{0.53\textwidth}
\setlength{\abovecaptionskip}{5pt plus 2pt minus 2pt}
\scriptsize
    \centering
        \includegraphics[width=0.53\textwidth]{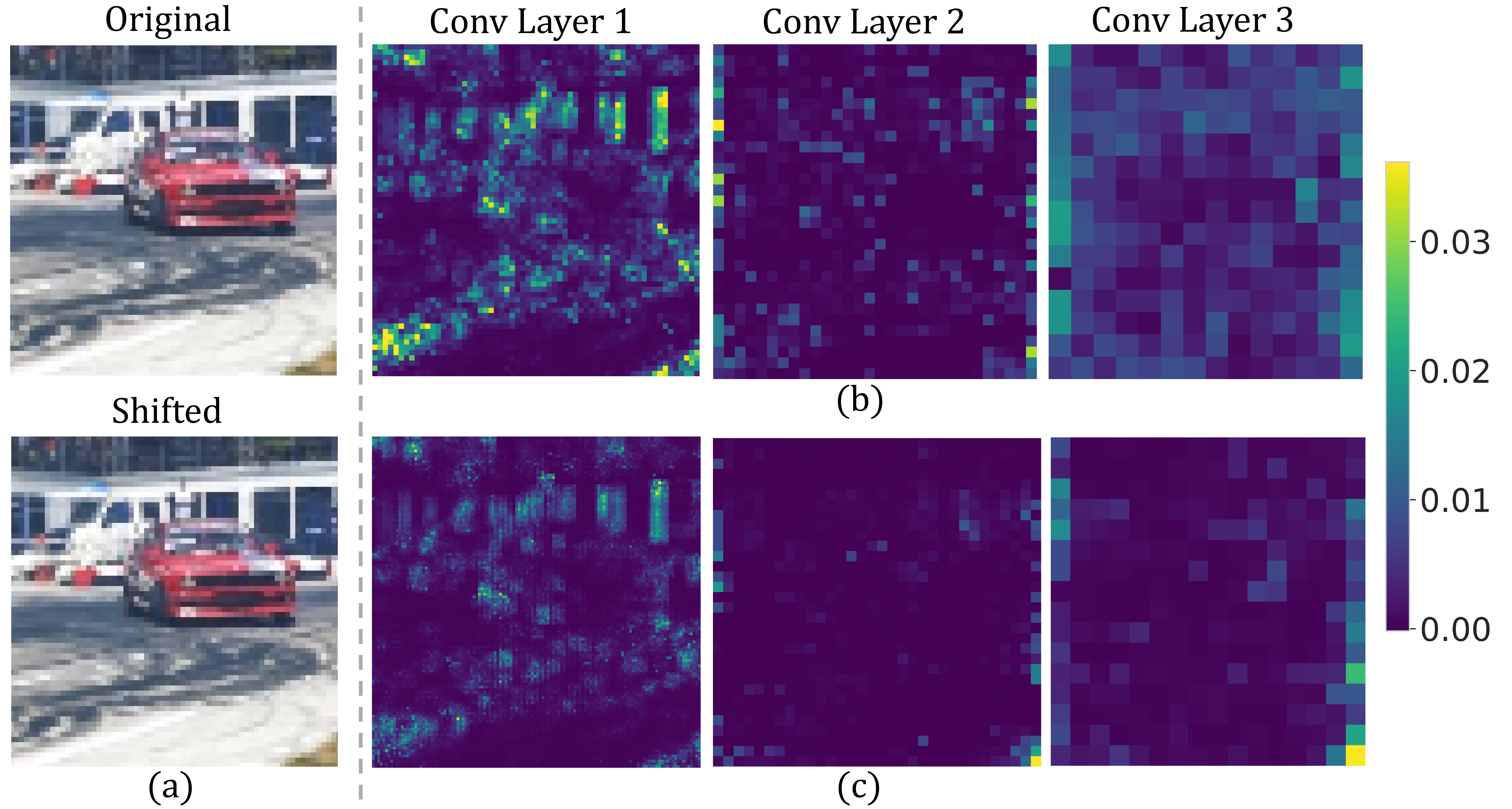}
    \caption{Feature embedding difference maps at different levels. (a) an input image and its one-pixel shifted version. (b) difference maps between embeddings extracted by the original AlexNet. (c) difference maps between embeddings extracted by AlexNet augmented with anti-aliased strided convolution and pooling layers.}
\label{fig:activationDifferences}
\end{wrapfigure}sensitive to small shifts as small shifts between two images $I_{dst}$ and $I_{ref}$ will be transferred to the shifts between their feature embeddings $F_{dst}$ and $F_{ref}$, which will in term drastically increase the distance between the feature embeddings $d(F_{dst}, F_{ref})$ as shown in Figure~\ref{fig:activationDifferences}(b). Downsampling in a neural network improves its shift invariance. Typically, downsampling can be achieved by a strided convolutional operator or a strided pooling operator with stride $n$ ($n > 1$). However, as discussed earlier in Reducing Stride\old{Section~\ref{sec:stride}}, striding introduces aliasing. While reducing stride size lessens aliasing, it prevents the network from reducing the feature size.

To keep the benefit of downsampling while reducing stride, \cite{zhang2019shiftinvar} invented a \emph{BlurPool} operator. Take max pooling with stride $n$ as an example. Such a max pooling operator can be decomposed into two steps: max pooling with stride 1, followed by a downsampling operator with stride $n$. To reduce the aliasing artifacts, \cite{zhang2019shiftinvar} followed the pre-filtering idea for anti-aliasing and replaced this max pooling operator with a sequence of three operators: a max pooling with stride 1, a Gaussian filter, and a downsampling operator with stride $n$. The last two operators are combined into as a single operator, called \emph{BlurPool}. Similarly, a convolution operator with stride $n$ can be replaced with its anti-aliased version as a convolution operator with stride $b$ and \emph{BlurPool} with stride $n/b$. \cite{zhang2019shiftinvar} found that replacing the original convolutional and pooling layers in a feature extraction neural network with their \emph{BlurPool} versions helps generate feature embeddings that make the downstreaming tasks more shift invariant. \emph{BlurPool} uses a fixed Gaussian filter for blurring and may lose some spatial features that are important attributes defining the quality of an image. \cite{zou2020delving} developed an adaptive anti-aliasing filter by learning a low-pass filter that is more content-aware. In this paper, we replace the strided convolution layers or pooling layers in the LPIPS framework with \emph{BlurPool} or adaptive anti-aliasing filters to make it invariant to imperceptible shifts among images in a pair. Figure~\ref{fig:activationDifferences} (c) shows that while anti-aliased convolution and pooling layers cannot make the feature network completely shift-invariant, they significantly reduce the difference between the feature embeddings from a pair of shifted images.

\textbf{Location of Anti-aliasing.} In a deep neural network, such as AlexNet used in LPIPS, a convolution layer is usually followed by an activation function. According to Zhang \cite{zhang2019shiftinvar}, the activation function is inserted between the stride-reduced convolutional layer and \emph{BlurPool}, as illustrated in Figure~\ref{fig:locationBP} (a). Vasconcelos \etal \cite{vasconcelos2021impact} created variants of the anti-aliased strided convolution by placing\begin{wrapfigure}{r}{0.5\textwidth}
\setlength{\abovecaptionskip}{5pt plus 2pt minus 2pt}
\scriptsize
\centering
    \includegraphics[width=0.5\textwidth]{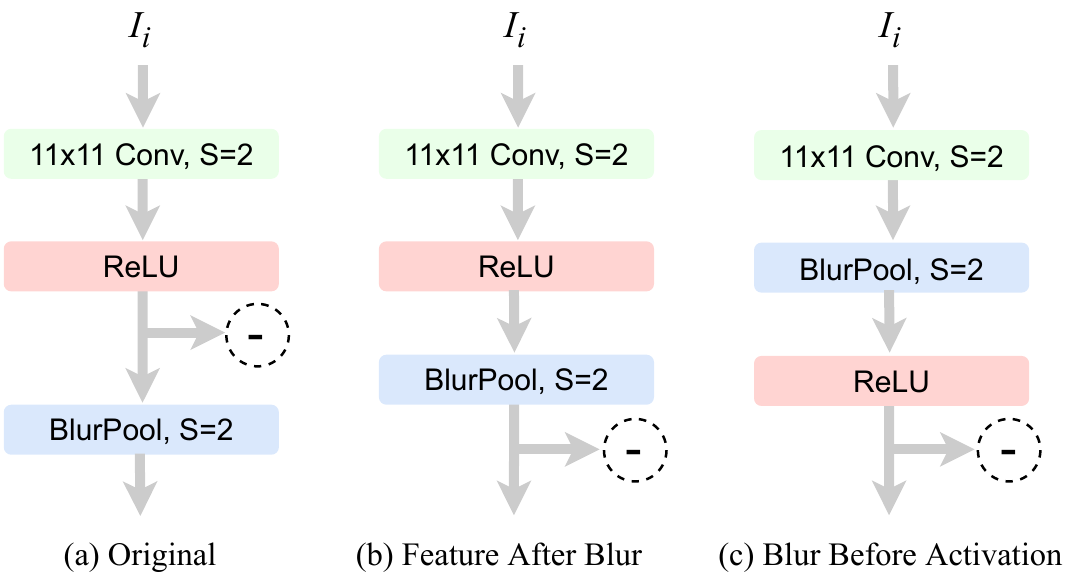}
\caption{Alternative positions of \emph{BlurPool}.}
\label{fig:locationBP}
\end{wrapfigure}the anti-aliasing filter at different locations, specifically, before or after the convolution operation. They found that some variants can lead to stronger learned inductive priors. But, will they provide significant improvements in shift tolerance? We build upon their findings and design variations of the anti-aliased strided convolutions. Specifically, we modify AlexNet \emph{conv-1} as illustrated in Figure~\ref{fig:locationBP} and explain the variants below.

\textit{Original.} As shown in Figure~\ref{fig:locationBP} (a), we follow the original design of \emph{BlurPool} and put it after \emph{ReLU}~\cite{zhang2019shiftinvar}. For anti-aliasing, the stride size of \emph{conv-1} is reduced from 4 to 2 and the \emph{BlurPool} layer has a stride of 2 so that the total stride of 4 is preserved in this anti-aliased version. We take the output of \emph{ReLU} as the feature embedding to calculate the similarity. 

\textit{Feature after blur.} In the above design, the feature embedding is used before \emph{BlurPool}. This effectively reduces the anti-aliasing effect on the feature embeddings although the reduced stride size in \emph{conv-1} still offers some level of anti-aliasing. Therefore, we investigated a variation of the anti-aliased convolution by taking the output of \emph{BlurPool} as the feature embedding to be used for similarity calculation, as illustrated in Figure~\ref{fig:locationBP} (b).

\textit{Blur before activation.} Vasconcelos \etal \cite{vasconcelos2021impact} suggested that blurring after the non-linearity, as done in Figure~\ref{fig:locationBP} (a) and (b), prevents high frequency from getting passed on to subsequent layers. Following their findings, we adopted their design by placing \emph{BlurPool} before \emph{ReLU} to keep the high-frequency information from \emph{ReLU}, as shown in Figure~\ref{fig:locationBP} (c).

\textbf{Border Handling.} Islam \etal reported that feature embeddings extracted by a CNN encode absolute position information~\cite{Islam*2020How}. This has an important implication for a learning-based similarity metric that feature embeddings from a CNN are position-dependent and are not shift-invariant. They found that zero padding can relieve this boundary problem for computer vision tasks that are sensitive to spatial information. Kayhan \& Gemert further proposed the concept of full convolution (F-Conv), in which every element of the filter needs to\setlength{\intextsep}{8pt}
\begin{wrapfigure}[11]{r}{0.44\textwidth}
\setlength{\abovecaptionskip}{8pt plus 2pt minus 2pt}
    \scriptsize
    \centering
    \includegraphics[width=0.44\textwidth]{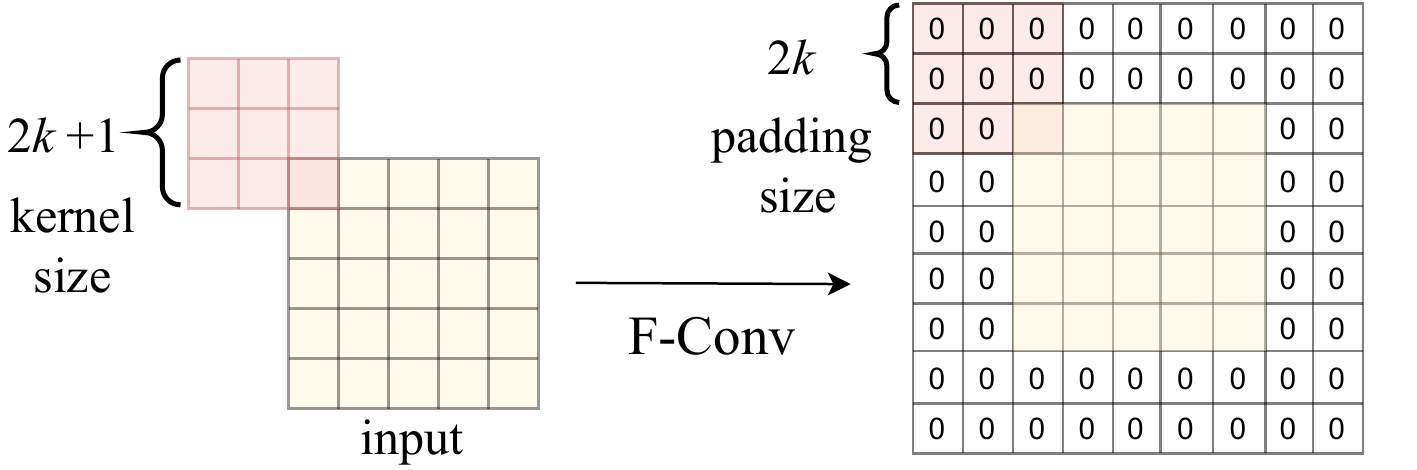}
    \caption{Full convolution applies every value of the filter to each value of an image. Hence, the input needs to be padded first with size 2$k$ for a filter with size $2k + 1$~\cite{Kayhan_2020_CVPR}.}
    \label{fig:fconv}
\end{wrapfigure}be applied to every pixel in the input image~\cite{Kayhan_2020_CVPR}. They implemented F-Conv as a regular convolutional operator with zero padding of $2k$ where $2k + 1$ is the filter kernel size as illustrated in Figure~\ref{fig:fconv}. Note, F-Conv will make the output of an un-strided convolution operator $2k$ larger than the input. They reported that F-Conv is least sensitive to the absolute position of the objects for image classification tasks. Inspired by these works, we replace the regular convolution operators with F-Conv in the LPIPS framework and increase the padding size in \emph{BlurPool} operators to achieve better shift-invariance.

\textbf{Pooling}. Max pooling is well known for being more shift invariant than average pooling. We investigate whether its anti-aliased version, \emph{MaxBlurPool} (described earlier in Anti-aliasing) is also more shift invariant than \emph{AvgBlurPool}, the anti-aliased version of average pooling when used in the LPIPS framework. Average pooling in its original form already supports anti-aliasing. We follow \cite{zhang2019shiftinvar} and implement \emph{AvgBlurPool} with a stride of $n$ as Gaussian filtering followed by \setlength{\intextsep}{0pt}
\begin{wrapfigure}[13]{r}{0.54\textwidth}
    \scriptsize
    \centering
    \includegraphics[width=0.53\textwidth]{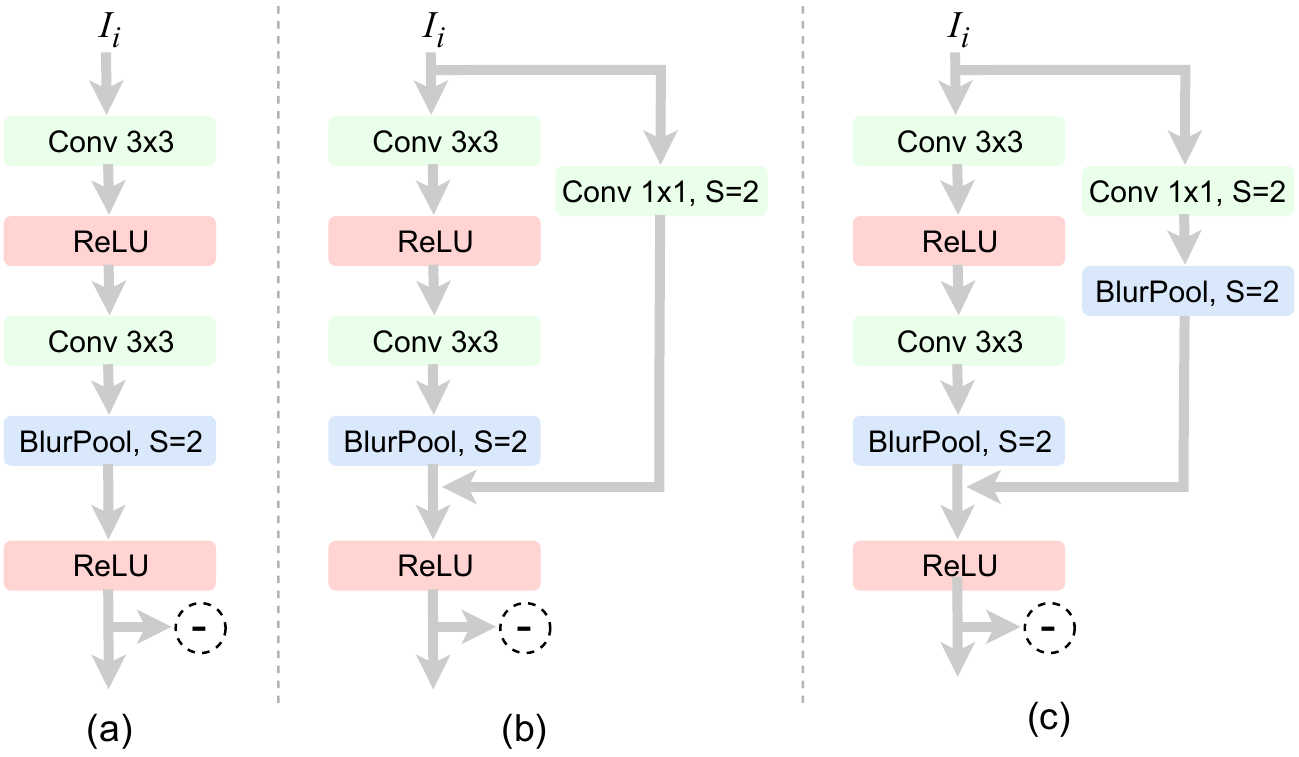}
    \caption{Anti-aliased skipped connection. (a) VGG-like network with AvgBlurPool, (b) with skip connection, and (c) with anti-aliased skip.}
    \label{fig:skip_archs}
\end{wrapfigure} downsampling with a factor of $n$.

\textbf{Strided-skip Connections.} Skip connection is widely used to speedup neural network training and obtain a high quality model. We investigate whether skip connection helps improve shift invariance of a learned similarity metric. As discussed in \cite{vasconcelos2021impact}, a strided skip connection introduces aliasing for the same reason discussed earlier in Anti-aliasing\old{Section~\ref{sec:aa}}. We therefore explore anti-aliased strided skip connections, as shown in Figure~\ref{fig:skip_archs}.

\section{Experiments} \label{sec:experiments}

We built upon the LPIPS framework and incorporated the elements discussed in Section~\ref{sec:methods} to investigate how these elements help develop a similarity metric that is consistent with the human visual similarity judgment and is robust against imperceptible shifts. We first compare our metrics to both off-the-shelf metrics, such as SSIM and MS-SSIM, and the recent learned similarity metrics. We then conduct ablation studies to evaluate how these elements work.

\textbf{Comparisons to Existing Metrics.} In Section~\ref{sec:benchmark}, we derived a shifted dataset from the BAPPS dataset and compared our metrics to existing metrics~\cite{sangnie2020,Ding20,Ma18:geo,kettunen2019lpips,Prashnani_2018_CVPR,wang2004image,wang2003multiscale,zhang2018perceptual}. In our experiments, we adopt the 2AFC score to evaluate how consistent a metric is with human judgment, and the rank-flipping rate, $r_{rf}$, to evaluate how robust it is against small shifts. As shown in Tables~\ref{tab:ti_study64} and \ref{tab:ti_study}, our metrics are both more consistent with human visual similarity judgment and more robust against imperceptible shifts than most of them, except a recent metric PIM~\cite{sangnie2020}, to which our method is comparable. PIM achieves shift robustness by training on neighboring video frames that often have small shifts. We work on an orthogonal solution by investigating neural network elements to make the learned metric robust and thus only train our metrics on the examples without any shift through data augmentation. We further evaluated the metrics on the perceptual validation dataset from the Challenge on Learned Image Compression~\cite{clic2021}. The results in Table~\ref{tab:clic} are consistent with previous results, i.e., our method outperforms\begin{wraptable}{r}{0.54\textwidth}
\setlength{\abovecaptionskip}{7pt plus 2pt minus 2pt}
\setlength{\belowcaptionskip}{-10pt plus 0pt minus 0pt}
    \begin{center}
    \scriptsize
    \caption{Experiments on the CLIC dataset.}
    \label{tab:clic}
    \addtolength{\tabcolsep}{-0.5pt}
    \begin{tabular}{lcccc}\toprule
        \multirow{3}{*}{Network} &\multirow{3}{*}{Accuracy(\%)} &\multicolumn{3}{c}{No. of rank flips} \\\cmidrule(lr){3-5}
        & &1pixel &2pixel &3pixel \\
        \midrule  
        L2 &58.16 &833 &2102 &2214 \\
        SSIM~\cite{wang2004image} &60.00 &349 &931 &1109 \\
        PIEAPP~\cite{Prashnani_2018_CVPR} &75.44 &91 &134 &158 \\
        E-LPIPS~\cite{kettunen2019lpips} &74.44 &212 &251 &317 \\
        DISTS~\cite{Ding20} &75.63 &28 &36 &50 \\
        PIM-1~\cite{sangnie2020} &73.79 &\textbf{13} &22 &33 \\
        LPIPS(Alex)~\cite{zhang2018perceptual} &73.68 &90 &108 &121 \\
        LPIPS(Alex)$^{\S * \dag}$ &76.53 &59 &51 &62 \\
        LPIPS(Alex) ours$^{* \dag}$ &\textbf{76.97} &17 &\textbf{14} &\textbf{21} \\
        \bottomrule
    \end{tabular}
    \end{center}
    \scriptsize{($\S$) Retrained from scratch. ($*$) Trained on image patches of size 64 using author's ($\dag$) setup.}
\end{wraptable}all others in terms of accuracy and is more shift-robust than others except PIM, which is similarly robust to ours.

\textbf{Ablation Studies.} We now examine how individual network elements affect our metrics. In these studies, we trained all our metrics using the original BAPPS training set on their original size of $256 \times 256$. We purposely did not train on the shifted version of the dataset to focus on neural network element designs. To train our metrics, we used the loss function: $MSE(s, h)$, where ${s = s_1/(s_1+s_2)}$, $s_1$ and $s_2$ are the predicted similarity scores of $I_1$ and $I_2$ to their corresponding reference, and $h$ is the human score. We trained our metrics using the same settings as \cite{zhang2018perceptual} except we used a lower dropout rate of 0.01. We tested all our metrics on the shifted dataset to obtain the rank-flipping rate. To obtain the 2AFC scores, we ran our metrics on full-size images (with no shift) of the original BAPPS dataset to verify whether our metrics sacrifice consistency with human visual similarity judgment to be robust against imperceptible shifts.

We first examine elements discussed in Section~\ref{sec:methods} individually. We use AlexNet as the backbone feature extraction network with the LPIPS framework as it provides the best result among other backbone networks~\cite{zhang2018perceptual}. As reported in Table~\ref{tab:alexnet_variants}, anti-aliasing via \emph{BlurPool} can greatly improve LPIPS's robustness against imperceptible shifts. Reducing stride size in its strided convolutional layer (\emph{conv-1}) also helps making it significantly more robust at little expense of the 2AFC score. Combining \emph{BlurPool} with reducing stride size makes the network even more robust against imperceptible shifts and more consistent with\begin{wraptable}[16]{r}{0.62\textwidth}
\setlength{\intextsep}{0pt}
\setlength{\abovecaptionskip}{7pt plus 2pt minus 2pt}
\setlength{\belowcaptionskip}{-5pt plus 0pt minus 0pt}
\newlength\wexp
\settowidth{\wexp}{Reflection-P}
\newcolumntype{C}{>{\centering\arraybackslash}p{\dimexpr.5\wexp-\tabcolsep}}
    \begin{center}
    \scriptsize
    \caption{Effect of (1) anti-aliasing (AA) via \emph{BlurPool}, (2) F-Conv, (3) reduced stride, \& (4) adaptive-AA$^\S$~\cite{zou2020delving} on learned metrics.}
    \label{tab:alexnet_variants}
        \begin{tabular}{CCcccccc}\toprule
            
            \multicolumn{2}{c}{AA (BlurPool)}
            &
            &
            &
            &
            \\
            
            \multicolumn{2}{c}{Reflection-Pad}
            &\multirow{2}{*}{F-Conv}
            &\multirow{2}{*}{Stride}
            &\multirow{2}{*}{2AFC} &\multicolumn{3}{c}{$r_{rf}$}
            \\
            \cmidrule(lr){1-2} \cmidrule(lr){6-8}
            
            1
            &2
            &
            & in \emph{conv-1}
            & 
            &1pixel
            &2pixel
            &3pixel
            \\
            \midrule  
            
            && &4 &70.65 &2.87 &3.92 &3.74 \\
            
            \checkmark & & &2 &70.53 &1.85 &2.22 &2.58 \\
            
            &\checkmark & &2 &\textbf{70.67} &1.46 &1.82 &2.25 \\
            
            \cdashline{1-8}\noalign{\vskip 0.6ex}
            
            &&\checkmark &4 &70.57 &2.78 &3.92 &3.91 \\
            &\checkmark &\checkmark &2 &70.52 &1.77 &2.15 &2.48 \\
            
            \cdashline{1-8}\noalign{\vskip 0.6ex}
            
            && &2 &70.54 &1.84 &2.28 &2.34 \\
            
            &\checkmark & &1 &70.42 &0.66 &\textbf{1.13} &1.83 \\
            
            &\checkmark &\checkmark &1 &70.44 &\textbf{0.63} &1.14 &\textbf{1.68} \\
            
            \cdashline{1-8}\noalign{\vskip 0.6ex}
            
            \checkmark$^\S$ & & &2 &70.57 &2.63 &3.36 &3.16 \\
             &\checkmark$^\S$ & &2 &70.63 &2.80 &3.57 &3.39 \\
            &\checkmark$^\S$ &\checkmark &2 &70.52 &2.95 &4.13 &3.93 \\
            
            \bottomrule
        \end{tabular}
    \end{center}
\end{wraptable}human judgment based on the 2AFC score. A larger reflection padding size also helps as it reduces the position information encoded in the feature embeddings from the image boundaries, as discussed in Section~\ref{sec:methods}. However, F-Conv, also designed to reduce the boundary issue, does not help. While the learned \emph{BlurPool}~\cite{zou2020delving} helps, it is not as effective as the original version for our task of making a robust similarity metric.

\setlength{\intextsep}{0pt}
\begin{wraptable}[16]{r}[0pt]{0.57\textwidth}
\setlength{\abovecaptionskip}{7pt plus 2pt minus 2pt}
\setlength{\belowcaptionskip}{-9pt plus 0pt minus 0pt}
\newlength\wrefl
\settowidth{\wrefl}{Reflection-P}
\newcolumntype{C}{>{\centering\arraybackslash}p{\dimexpr.5\wrefl-\tabcolsep}}
    \begin{center}
    \scriptsize
    \caption{Anti-aliasing via \emph{BlurPool} can significantly improve shift-tolerance and often improve 2AFC scores consistently for different backbone networks.}
    \label{tab:lpips_nets}
        \begin{tabular}{cCCcccc}\toprule
            \multirow{3}{*}{Network} 
            &\multicolumn{2}{c}{AA (BlurPool)}
            &\multirow{3}{*}{2AFC} 
            &
            \\
            
            &\multicolumn{2}{c}{Reflection-Pad}
            & 
            &\multicolumn{3}{c}{$r_{rf}$}
            \\
            \cmidrule(lr){2-3} \cmidrule(lr){5-7}
            
            &1
            &2
            &
            &1pixel
            &2pixel
            &3pixel
            \\
            \midrule 
            
            \textbf{VGG-16} & & &70.03 &3.01 &3.76 &3.44 \\
            &\checkmark & &70.05 &\textbf{0.66} &\textbf{1.08} &\textbf{1.44}  \\
            & &\checkmark &\textbf{70.07} &\textbf{0.66} &1.12 &1.82  \\
            
            \cdashline{1-7}\noalign{\vskip 0.6ex}
            
            \textbf{ResNet-18} & & &69.86 &2.67 &3.35 &3.77  \\
            &\checkmark & &69.95 &\textbf{0.82} &\textbf{1.51} &\textbf{2.19}\\
            & &\checkmark &\textbf{70.14} &1.07 &1.81 &2.38  \\
            
            \cdashline{1-7}\noalign{\vskip 0.6ex}
            
            \textbf{Squeeze} & & &69.61 &7.41 &7.58 &10.35 \\
            &\checkmark & &69.24 &\textbf{2.03} &3.06 &3.93 \\
            & &\checkmark &\textbf{69.44} &2.10 &\textbf{2.48} &\textbf{3.42}  \\
            
            \bottomrule
        \end{tabular}
    \end{center}
\end{wraptable}

We test on different backbone feature networks, including VGG-16~\cite{simonyan2014very}, ResNet-18~\cite{He_CVPR_2016}, and SqueezeNet~\cite{SqueezeNet}. While reducing the stride size is effective, not all networks have a strided convolution layer. Hence, we focus on \emph{BlurPool} applied to pooling layers. As shown in Table~\ref{tab:lpips_nets}, \emph{BlurPool} significantly improves the robustness of other backbone networks as well. What is interesting is the effect of the padding size within these backbone networks. While a larger padding size improves 2AFC, it does not make shift-invariance better for VGG-16 and ResNet-18. For them, the anti-aliased \emph{conv} layers have a stride size of 1, which leads to minor boundary issues. We conjecture that this makes a larger padding size unnecessary.

We also examine the effect of the location of \emph{BlurPool} within AlexNet. As reported in Table~\ref{tab:location_blurpool}, the original version (Figure~\ref{fig:locationBP} (a)) works best when the stride size is 2 in \emph{conv-1}. With a smaller stride size, it does not work as well as Blur Before Activation (Figure~\ref{fig:locationBP} (c)). This is in part consistent with what was found by Vasconcelos \etal \cite{vasconcelos2021impact}. In the original design, \emph{BlurPool} is placed after the activation layer for anti-aliasing at the expense of the reduction of the high-frequency information from the activation layer. With the need for anti-aliasing due to a larger stride size, this trade-off works out. However, when stride size is 1, the need for anti-aliasing is reduced; therefore, it is more helpful to place \emph{BlurPool} before the activation layer to avoid the loss of high-frequency information. Thus, Blur Before Activation works better when the stride size\begin{wraptable}[11]{r}{0.68\textwidth}
\setlength{\abovecaptionskip}{7pt plus 2pt minus 2pt}
\setlength{\belowcaptionskip}{0pt plus 0pt minus 0pt}
    \begin{center}
    \scriptsize
    \caption{Effect of \emph{BlurPool} locations within an anti-aliased strided convolution  (Figure~\ref{fig:locationBP}).}
    \label{tab:location_blurpool}
        \begin{tabular}{ccccccc}\toprule
            
            \multirow{2}{*}{Anti-Alias}
            &\multirow{2}{*}{Stride}
            &\multirow{2}{*}{BlurPool}
            &\multirow{2}{*}{2AFC} &\multicolumn{3}{c}{$r_{rf}$}
            \\
            \cmidrule(lr){5-7}
            
            (BlurPool)
            & in \emph{Conv-1}
            & Location
            & 
            &1pixel 
            &2pixel 
            &3pixel 
            \\
            
            \midrule 
            
            \checkmark &2 &Original &\textbf{70.67} &\textbf{1.46} &\textbf{1.82} &\textbf{2.25}  \\
            \checkmark &2 &FeatAfterBlur &70.55 &1.73 &1.84 &2.49  \\
            
            \checkmark &2 &BlurBeforeAct &70.50 &2.06 &2.02 &2.74 \\
            
            \cdashline{1-7}\noalign{\vskip 0.6ex}
            
            \checkmark &1 &Original &70.42 &0.66 &1.13 &1.83  \\
            
            \checkmark &1 &FeatAfterBlur &\textbf{70.52} &0.69 &1.11 &1.60  \\
            
            \checkmark &1 &BlurBeforeAct &70.48 & \textbf{0.57} &\textbf{1.06} &\textbf{1.50} \\
            
            \bottomrule
        \end{tabular}
    \end{center}
\end{wraptable}is 1. We also observed that \emph{MaxBlurPool} has better shift tolerance but lower 2AFC scores (accuracy) than \emph{AvgBlurBool}. 
Moreover, using anti-aliased strided-skip connections leads to higher accuracy with a negligible drop in shift tolerance.

\newpage
\begin{wraptable}[17]{r}{0.38\textwidth}
\setlength{\abovecaptionskip}{7pt plus 2pt minus 2pt}
\setlength{\belowcaptionskip}{-10pt plus 0pt minus 0pt}
    \begin{center}
    \scriptsize
    \caption{Consistency of perceptual similarity metrics with the sensitivity of human perception to pixel shifts.}
    \label{tab:jnd}
        \begin{tabular}{lc} \toprule
            Metric
            & JND mAP\%
            \\
            \midrule
            SSIM~\cite{wang2004image} &0.722 \\
            LPIPS (Alex)~\cite{zhang2018perceptual} &0.757 \\
            LPIPS (Alex) $^{\S * \dag}$ &0.740 \\
            LPIPS (Alex) \textbf{ours}$^{* \dag}$ &0.771 \\
            LPIPS (VGG)~\cite{zhang2018perceptual} &0.770 \\
            LPIPS (VGG) $^{\S * \dag}$ &0.769 \\
            LPIPS (VGG) \textbf{ours}$^{* \dag}$ &\textbf{0.775} \\
            DISTS~\cite{Ding20} & 0.766 \\
            PIM-1~\cite{sangnie2020} &0.773 \\
        \bottomrule \end{tabular} \\
    \end{center}
    \scriptsize{($\S$) Retrained from scratch. ($*$) Trained on image patches of size 64 using author's ($\dag$) setup.}
\end{wraptable}

\textbf{Just noticeable differences (JND).} We conducted the following experiment to study how consistent our shift-tolerant perceptual similarity metric is with the human perception results reported in Table~\ref{tab:pixelShifts}. In our study reported in Table~\ref{tab:pixelShifts}, we had asked our participants if the two images, which may be shifted by a few pixels, were the same or not. Using these responses, we perform a just noticeable difference test. We use only those samples which have at least 3 human responses. There were 301 such samples, and the mean number of samples per pixel-shift (0 to 9) is 30.1 with a standard deviation of 1.6 (maximum 33 and minimum 28). Following Zhang \etal ~\cite{zhang2018perceptual}, we rank the pairs by a perceptual similarity metric and compute the area under the precision/recall curve (mAP)~\cite{everingham2010pascal,zhang2018perceptual}. The results in Table~\ref{tab:jnd} show that our shift-tolerant LPIPS metrics follow the sensitivity of human perception to pixel shifts more accurately than their vanilla versions. The accuracy of PIM-1 and DISTS is comparable to ours.

\textbf{Summary.} Among the network elements we investigated, anti-aliased strided convolution, anti-aliased pooling, and reduction of stride size are most effective to develop a perceptual similarity metric that is robust against imperceptible shifts. These findings are consistent across a variety of backbone network architectures. A larger padding size helps reduce the position information due to the boundary issues encoded in the feature embeddings. Anti-aliased skip connection can help improve accuracy but with little effect on shift invariance. The position of \emph{BlurPool} matters. It should be placed before the activation layer if its precedent convolution uses a small stride size.

\section{Conclusion}

This paper reported our investigation on how to design a deep neural network as a learned perceptual image similarity metric that is both consistent with the human visual similarity judgment and robust against the imperceptible shift among a pair of images. We discussed various neural network elements, such as anti-aliased strided convolution, anti-aliased pooling, the placement of \emph{BlurPool}, stride size, and skip connection, and studied their effect on a similarity metric. We found that using anti-aliasing strided convolutions and pooling operators and reducing stride size help to make a learned similarity metric shift-invariant. Our experiments show that by integrating these elements into a neural network, we are able to develop a learned metric that is more robust against imperceptible shifts and more consistent with the human visual similarity judgment.

\noindent\textbf{Acknowledgments.} Figure~\ref{fig:teaser} uses frames from  \url{https://www.youtube.com/watch?v=jW7pFhkVNYY} under 
a Creative Commons license.

\clearpage

\bibliographystyle{splncs04}
\bibliography{egbib}

\end{document}